\pdfoutput=1
\documentclass[11pt]{article}

\usepackage[preprint]{acl}

\usepackage{listings}
\usepackage{xcolor}
\usepackage{times}
\usepackage{latexsym}
\usepackage{enumitem}
\usepackage{colortbl} % 彩色表格
\usepackage{tcolorbox}
\usepackage{multirow}
\usepackage{booktabs}
\usepackage{amsmath}
\usepackage[T1]{fontenc}
\usepackage[utf8]{inputenc}

\usepackage{microtype}

\usepackage{inconsolata}

\usepackage{graphicx}
\tcbuselibrary{listings,theorems,breakable}
\newtcbtheorem[number within=section]{exmp}{Case}%
{breakable,colback=white!5!white,colframe=black!95!,fonttitle=\bfseries, left=.02in, right=.02in,bottom=.02in, top=.02in}{exmp}

\title{Teaching LLM to Reason: Reinforcement Learning from Algorithmic Problems without Code}

\author{Keqin Bao\textsuperscript{1,2},
        Nuo  Chen\textsuperscript{2,3},
        Xiaoyuan Li\textsuperscript{1,2}\thanks{Work done when Keqin Bao, Nuo Chen, and Xiaoyuan Li are intern at Alibaba Group.},  
        Binyuan Hui\textsuperscript{2} \\
        \textbf{Bowen Yu\textsuperscript{2}},
        \textbf{Fuli Feng\textsuperscript{1}},
        \textbf{Xiangnan He\textsuperscript{1}},
        \textbf{Dayiheng Liu\textsuperscript{2}} \\
        University of Science and Technology of China\textsuperscript{1} \\
        Alibaba Group\textsuperscript{2} \\
        The Hong Kong
University of Science and Technology (Guangzhou)
\textsuperscript{3} \\
        }

\begin{document}
\maketitle
\begin{abstract}
    % This is an abstarct

Enhancing reasoning capabilities remains a central focus in the LLM reasearch community. A promising direction involves requiring models to simulate code execution step-by-step to derive outputs for given inputs. However, as code is often designed for large-scale systems, direct application leads to over-reliance on complex data structures and algorithms, even for simple cases, resulting in overfitting to algorithmic patterns rather than core reasoning structures.
To address this, we propose \textbf{TeaR}, which aims at \textbf{Tea}ching LLMs to \textbf{R}eason better. TeaR leverages careful data curation and reinforcement learning to guide models in discovering optimal reasoning paths through code-related tasks, thereby improving general reasoning abilities. We conduct extensive experiments using two base models and three long-CoT distillation models, with model sizes ranging from 1.5 billion to 32 billion parameters, and across 17 benchmarks spanning Math, Knowledge, Code, and Logical Reasoning. The results consistently show significant performance improvements. Notably, TeaR achieves a 35.9\% improvement on Qwen2.5-7B and 5.9\% on R1-Distilled-7B .
\end{abstract}

\section{Introduction}

With the development of Reasoning-Enhanced Large Language Models (Reasoning-Enhanced LLMs) (e.g., o1~\cite{jaech2024openai} and R1~\cite{guo2025deepseek}), they have demonstrated significant potential in solving complex tasks~\cite{DBLP:journals/corr/abs-2503-09567,li2025mtr}. Current research and applications mostly focus on improving the reasoning performance of models in specific domains such as mathematics~\cite{shao2024deepseekmath,yang2024qwen25mathtechnicalreportmathematical,Toshniwal2024OpenMathInstruct1A1,li2024evaluating}, science~\cite{DBLP:conf/nips/LuMX0CZTCK22,rein2024gpqa,DBLP:journals/corr/abs-2502-14739}, commonsense~\cite{dua-etal-2019-drop, li2025hellaswag}, or code generation ~\cite{hui2024qwen25codertechnicalreport,jain2024livecodebench,guo2024deepseekcoderlargelanguagemodel}. However, the truly valuable enhancement should be a broad improvement in reasoning capabilities, rather than being confined to a specific domain, which might be a pathway towards AGI.

Existing research highlights code as a key element for enhancing reasoning abilities~\cite{wang2023mathcoder, li2025codeiocondensingreasoningpatterns, ding2024semcodertrainingcodelanguage}, due to its complex reasoning patterns and ability to solve real-world problems through step-by-step operations. There are currently two main approaches to leveraging code data to enhance reasoning abilities: one is directly training on raw code or code-text pairs~\cite{gong2025pseudocode,hui2024qwen25codertechnicalreport}, while the other involves requiring the model to infer the outputs of the test case inputs based on the logic of the code’s reasoniing pattern, and synthesizing data using rejection sampling~\cite{DBLP:journals/corr/abs-2308-01825}. Despite these efforts achieving notable success, they still struggle with several limitations.

Regarding the former approach, it often struggles to extract clear reasoning signals due to their implicit nature and entanglement with syntactic noise. Even structured objectives like text-to-code generation remain limited by their dependence on code-specific syntax, hindering the transfer of learned representations to broader, non-code tasks \cite{li2025codeiocondensingreasoningpatterns}. 
For the later, it encourages reasoning via execution-based supervision and synthesized problem-solving, but overlooks that code is often optimized for large-scale data. This leads to the use of complex algorithms and data structures that are unnecessary for small test case inputs. Enforcing such optimizations risks overfitting models to algorithmic patterns rather than general reasoning ability (See Section \S\ref{sec:2.2} for more details). 

\begin{figure*}[t]   
\centering
\setlength{\abovecaptionskip}{-0.10cm}
\setlength{\belowcaptionskip}{0cm}
\includegraphics[width=0.85\linewidth, height=0.25\textheight,scale=0.9]{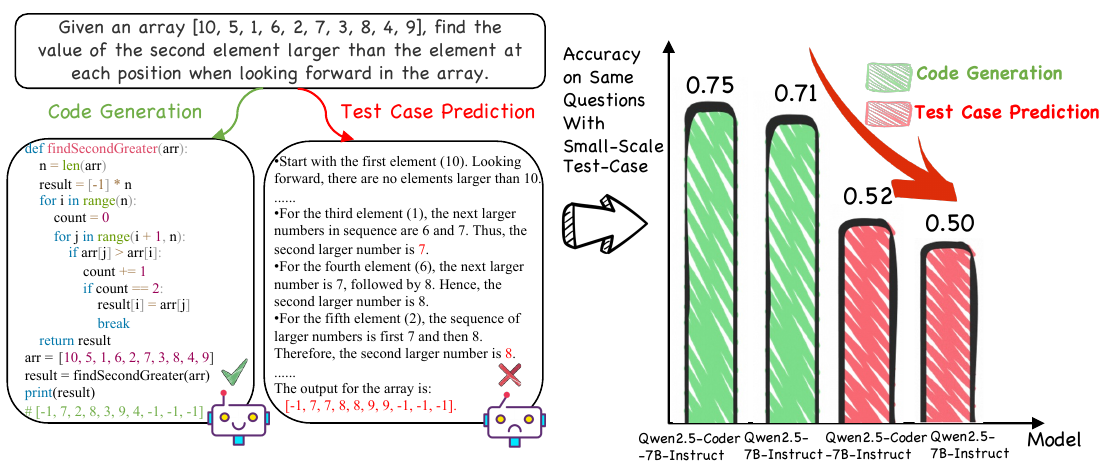}
\caption{The right shows the difference in code generation and test case prediction accuracy between Qwen2.5-7B-Instruct and Qwen2.5-Coder-7B-Instruct for the same queries. The left figure gives a tiny example.}
\label{example}
\vspace{-3pt}
\end{figure*}
% Based on these considerations, we argue that the key to enhancing reasoning lies not merely in exposing models to code logic, but in engaging them with tasks that inherently demand logical deduction. While code can serve as a valuable source of structured reasoning, its direct usage may introduce unnecessary algorithmic patterns that obscures fundamental reasoning processes. Instead, we propose empowering models to tackle such tasks through their own reasoning capabilities. Instead, our preliminary experiments find that allowing models to explore reasoning methods on their own for code-related tasks is also a powerful way to improve general reasoning.
Based on these considerations, we argue that improving reasoning requires engaging models in tasks that demand logical deduction, rather than just exposing them to code logic. While code offers structured reasoning, its direct use may introduce complex algorithmic patterns that obscure core reasoning. Instead, empowering models to explore reasoning on their own may be more effective for developing general reasoning skills.
To achieve this, inspired by the idea that Reinforcement Learning (RL) can elicit latent reasoning patterns within base models~\cite{yue2025does}, we explore RL as a way to elicit reasoning skills. 

% Specifically, we first perform careful data selection, including executability filtering, test-case complexity filtering, and then adopt GRPO to train models to infer test case outputs from inputs by autonomously searching for suitable reasoning paths through Chain-of-Thought (CoT). This approach encourages models to develop and refine their own reasoning patterns, fostering deeper understanding and more flexible application of reasoning across diverse domains.
Specifically, we propose \textbf{TeaR}, a simple yet effective method aims to \textbf{Tea}ch LLM to \textbf{R}eason better. To acheive this, we begin with careful data selection for code-related tasks via algorithmic problems without using code, filtering out easily guessable problems (e.g., ``yes''/``no'' answers) through rule-based and model-based strategies. We further filter instances based on code executability, runtime constraints, and test case complexity. Considering the high demand of algorithmic tasks on reasoning ability, we use algorithmic tasks as our task source. In total, we collect 200K queries from 2.4K distinct algorithmic tasks. We then employ GRPO~\cite{shao2024deepseekmath} to train models to infer test case outputs from inputs by autonomously discovering effective reasoning paths. Leveraging the deterministic nature of code-related outputs, we use a rule-based reward based on exact match for training. This approach encourages models to develop and refine their own reasoning strategies, promoting deeper understanding and greater adaptability across diverse reasoning tasks.

% Through in-depth experiments and anlysis covers training from base model (zero setting) and training from long-cont distilled models, 我们发现通过模型自己在reasoning 强相关的任务探索合适的reasoning pattern

Through extensive experiments and analysis—including training from a base model and from long-CoT distilled models across scales from 1.5B to 32B—we evaluate TeaR on 17 public datasets spanning mathmatics, knowledge-intensive, code, and logical reasoning. The results demonstrate that TeaR consistently improves performance across all domains. Our key findings are as follows: (1) TeaR shows more efficient task utilization, even outperforms the strong baseline Code I/O~\cite{li2025codeiocondensingreasoningpatterns} on the mathmatics tasks with only 2.4k distinct seed queries.  (2) Even for distilled models that are heavily optimized for math/code tasks, TeaR still brings performance gains, highlighting its broad applicability. (3) In-depth analysis shows that TeaR exhibits strong generalization in reasoning abilities, even improving performance on entirely unrelated reasoning tasks, such as user behavior reasoning for movie recommendations.

In summary, our contributions are as follows:
\begin{itemize}[leftmargin=*]
    \item We highlight that leveraging code for reasoning should focus on the task itself, rather than merely imitating code execution logic.
    \item We propose TeaR, which directly incorporates code-related tasks but avoids using code data in RL, effectively enhancing general reasoning capabilities.
    \item Extensive experiments on 17 public datasets demonstrate the effectiveness of our approach.
\end{itemize}

% code的使用场景往往是为了大规模的数据和系统设计的，这也要求了其

% 其富有逻辑性的代码关系此外we 们认为 not only code  but also code related task 是reasoning强相关的，这些task需要缜密的逻辑来进行推理
% 现在的使用code的方式 either 以来训练coding代码 code text对 或者利用reference code code <--> text 之间存在mismatch  （最后
% mismatch 是什么？mismatch 是 对于简单的test case input来说 即使是专业的算法工作者也不会使用负责的算法来进行解决，精妙的算法往往是为了大型/复杂的系统设计的，例如 我们在解决小规模的问题是不需要借助复杂的数据结构，可以接受O(n^2)甚至O(n^3)的计算复杂度当时当数据规模大到一定程度时我们往往需要数据结构进行优化，因此在简单的test case input上使用复杂数据结构进行参考反而会舍近求远,同样可能限制模型的探索空间和使用自身能力进行求解的能力

% In this paper, three motivations:
% (1) It is not merely the code itself, but rather the tasks associated with code that require rigorous logical reasoning.
% (2) Directly using code may not be the optimal solution.
% (3) Instead, we should focus on empowering models to explore these tasks with strong reasoning demands through their own capabilities.
%   在这篇文章中，
% （1）不仅仅是代码 是代码相关的任务需要缜密的逻辑推理 （2） 直接使用代码可能不是最优解 (3) 我们应当

% 在众多推理相关任务中，We believe that real-world code programs reflect the integration of a wide range of reasoning patterns across diverse
% contexts,  这是因为与代码相关的任务往往需要模型具备严密的逻辑推导能力。因此，研究者们认为代码及其相关任务是训练和评估模型推理能力的理想媒介。

% 然而，目前大多数方法在利用代码数据时存在一定的局限性。现有做法通常依赖于“代码-文本”对进行训练，或通过参考代码作为解题思路的来源。但这种方式在实际应用中常常面临一个关键问题： 。例如，在面对简单测试用例或输入时，即使是专业的算法工程师也不会选择使用复杂的数据结构或高级算法来求解。事实上，许多精妙的算法设计初衷是为了应对大规模或高复杂度的问题场景。对于小规模任务而言，接受较高时间复杂度的解决方案是合理且高效的。若在此类场景下强制引入复杂的算法结构，反而可能限制模型自身的探索空间，并阻碍其发挥出更自然、直接的解题能力。

% 因此，在本文中，我们提出一种新的视角与方法，旨在更有效地利用代码相关任务来全面提升大语言模型的推理能力，同时避免传统方法中存在的语义错配问题。我们将进一步探讨如何构建更具代表性和泛化性的训练信号，以引导模型在不同推理场景下都能表现出更强的理解与逻辑推导能力。
\section{Preliminary}

In this section, we will elaborate on our observations. First, we will explain that learning to solve a task with code doesn't imply understanding it through logical reasoning (\S\ref{sec:2.1}). Then, we will provide examples to demonstrate that requiring a model to follow the reasoning process from code-execution might not be a good solution for all cases (\S\ref{sec:2.2}).
\subsection{From Code to Task}
\label{sec:2.1}
%
% 1. 学会用代码解决任务 不代表学会这个任务
The pursuit of reasoning capabilities in LLMs has long been intertwined with code-related tasks. Traditional approaches~\cite{DBLP:journals/pacmpl/DingMKR24,DBLP:conf/nips/DingPMKYR24} focus mainly on code generation, where models are trained to learn programming syntax, algorithmic structures, and software engineering paradigms. 
The underlying hypothesis posits that exposure to code's intricate logic patterns will implicitly cultivate generalizable reasoning abilities. However, as previous studies have pointed out, reasoning patterns in code are expressed through highly structured syntax. Directly learning these implicit patterns may cause models to overfit to code-related reasoning tasks, limiting their ability to generalize to other domains or tasks.

We argue that the true bottleneck lies in the misalignment between training objectives and desired capabilities.
Code, in essence, is a process of transforming human logic, ideas, and problem-solving approaches into language that a computer can execute. What we need to learn might not be the syntax and commands, but rather how the task is solved -- the core reasoning ability.

To validate this, we conduct preliminary experiments by evaluating the model on a random sample of LeetCode easy-level problems involving both code generation and test case prediction. Example and results are shown in Figure~\ref{example}. 
% Our key finding is that, although Qwen2.5-Coder-7B-Instruct achieves 75\% accuracy on code generation, it only reaches 52\% on simplified test case prediction tasks requiring equivalent logical reasoning.
We observe that even on very simple tasks, the models are capable of performing code generation but struggle with text-based reasoning. Statistically, there is an approximate 20\% performance gap between code generation and test case prediction (i.e., small-scale test cases) across both models.
This performance gap contradicts human cognition—software engineers who can write the code rarely fail on such reasoning tasks.
This suggests that current models conflate reasoning patterns with code-specific representations, highlights the need to focus on \textbf{learning the task}. 
% Thus, current models fail to decouple reasoning patterns from their code-specific representations. This findings encourages us to shift attention towards the core reasoning tasks, rather than being confined to their particular expression in the form of code generation.

\subsection{Code is Not Always We Need}
\label{sec:2.2}
In practical programming, code is often designed to handle large-scale data, making efficiency issues more important. 
% For example, when dealing with massive datasets, we typically employ efficient data structures and algorithms like hash tables, segment trees, and dynamic programming to ensure that programs can complete tasks within limited time and space. 
However, when faced with small-scale inputs, these ``optimal solutions'' may not be the best choice from the perspective of natural language reasoning.

An example is shown in Appendix~\ref{app:case}. Suppose we need to frequently perform addition and find the maximum value over subintervals of an array. In large-scale data scenarios, we would use advanced data structures to optimize the algorithm's time complexity to O($n logn$). However, if the input array contains only a few elements and the number of queries is very limited, following the same approach to reason could lead to generating excessive tokens and deviating from intuitive human reasoning. As illustrated in the Appendix~\ref{app:case}, strictly following the code-based execution would require 695 tokens for reasoning, while allowing the model to explore the solution on its own also leads to the correct result but with only 236 tokens. 
Taking this a step further, if we train LLMs to completely learn and reproduce these optimized coding logics even in cases where such optimization is unnecessary. we risk producing overly long token sequences or overfitting to specific algorithmic patterns. Therefore, it might be better to let the model explore the most effective and suitable solution of the question by itself to improve its general reasoning capabilities.
Then we will focus on how to utilize algorithmic tasks without code to enhance the model's general reasoning capabilities.

% 2. case study  
\section{TeaR}
\begin{figure*}[t]   
\centering
\setlength{\abovecaptionskip}{-0.10cm}
\setlength{\belowcaptionskip}{0cm}
\includegraphics[width=0.9\linewidth,scale=0.9]{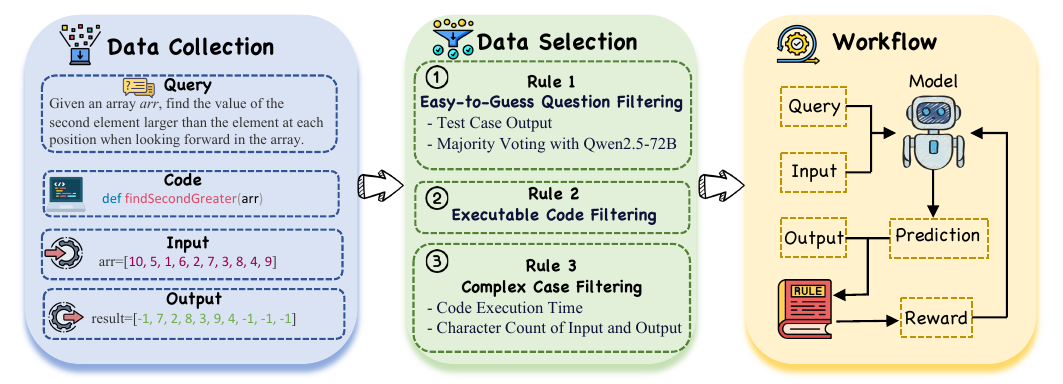}
\caption{The overview of our TeaR framework}
\label{overview}
\vspace{-5pt}
\end{figure*}
% 在这一章节中，我们首先将结合RL的训练方式说明述为什么我们使用RL，然后我们将讲述我们怎么对数据进行选取。
In this section, we first elaborate on why we adopt reinforcement learning by discussing our workflow (\S\ref{sec:3.1}). Following that, we will describe how we select the data for our experiments (\S\ref{sec:3.2}), our training pipeline is illustrated in Figure~\ref{overview}.
\subsection{Workflow}
\label{sec:3.1}
Building upon the observations discussed in the preceding sections, we posit that traditional supervised learning paradigms may not fully harness the reasoning capabilities inherent to LLMs. Generally, imposing predefined code-based reasoning patterns on the model can result in suboptimal performance, often forcing it to rely on unnecessarily complex strategies. 
Instead, we believe that empowering models to autonomously explore and identify the most effective reasoning behaviors from their own perspective could lead to more efficient and robust solutions. 
This perspective resonates closely with the fundamental principle of reinforcement learning, which emphasizes adaptive, goal-oriented \textbf{exploration}~\cite{simchowitz2024exploration}. Unlike supervised learning, which primarily focuses on replicating patterns, RL enables models to learn based on desired outcomes, thereby activating reasoning strategies that have already been internalized during pretraining~\cite{chu2025sft}. 

To tap into this self-activated reasoning potential, we employ GRPO, an advanced RL algorithm tailored specifically for LLMs. In detail, during the training phase, we provide the model with an algorithmic question and its corresponding correct output, allowing the model to explore the solution space autonomously. Thanks to the deterministic of executable code outputs, we can use extact match as our reward metric, which provides a clear and unambiguous evaluation signal. Accordingly, our reward function is defined as:
$$
\small
R(x, y) = 
\begin{cases} 
1, & \text{if } y_{\text{pred}} = y_{\text{true}}, \\
0, & \text{otherwise}.
\end{cases}
$$

\subsection{Data Selection}
\label{sec:3.2}
To ensure the quality and usability of our dataset, we adopt a systematic four-step filtering process that emphasizes accuracy, efficiency, and learnability. This methodology is designed to produce a dataset that is both reliable for natural language reasoning tasks and practical for model training. We adopt LeetcodeDataset~\cite{xia2025leetcodedataset} as the seed data and apply the following processing pipeline.
\begin{itemize}[leftmargin=*]
    \item \textbf{Easy-to-Guess Question Filtering} To prevent the model from easily guessing the answers (e.g. ``Yes''/``No'', ``Alice''/``Bob'') to given questions during RL learning~\cite{eisenstein2024helping}, we first excluded some easily guessable questions from the code tasks. To achieve this, we employed two methods: one involves filtering based on test case outputs, directly removing questions with potential answers less than or equal to 8; the other requires Qwen2.5-72B to perform majority voting to determine whether each question is likely to be guessed, and questions deemed easy to guess are removed.
    \item \textbf{Executable Code Filtering} To ensure problem accuracy and input-output consistency, we validate each code sample to confirm that it is executable and capable of producing the correct output for all corresponding test case inputs.
    \item \textbf{Complex Case Filtering} For overly complex test cases (e.g., inputs that are too large, containing more than 1000 numbers), reasoning through such samples not only consumes a large number of tokens, but also poses significant challenges for most models. Therefore, it is very difficult to directly ask models to perform reasoning on such complex cases. As a result, we apply rule-based filtering to remove relatively complex test cases. This includes two rules: (1) Filtering based on code execution time, due to instability in code compilation, execution, and sandbox environments, we manually examined a large number of cases and decided to remove samples where the execution time exceeds 1 second.   (2) Filtering based on the character count of input and output, handling excessively long input or output texts tests the model's ability to understand and recall long contexts. Therefore, we remove test cases where either the input or output exceeds 200 characters.
    % To 
    % The first step involves verifying the executability of all code snippets in the dataset. Each snippet is tested in a controlled environment to ensure it runs without errors and produces outputs consistent with its inputs. Any code that fails to execute or exhibits mismatched input-output pairs is excluded.

\end{itemize}
\vspace{-2pt}
After a careful curation process, we constructed a training set consisting of 200K examples based on 2.4K seed questions.

% \textbf{Rationale}: Ensuring executability guarantees the dataset's reliability, enabling downstream tasks to focus on reasoning rather than error correction.

% \textbf{Rationale}: Removing inefficient code reduces computational overhead and ensures that the dataset remains focused on tasks that align with the capabilities of modern language models.

% \textbf{Input/Output Length Constraints}
% To maintain simplicity and focus, we impose strict character limits on both input and output fields, capping each at 200 characters. Data entries exceeding this limit are discarded.

% \textbf{Rationale}: Excessively long inputs or outputs hinder natural language reasoning by introducing unnecessary complexity. Shorter data entries allow models to focus on core reasoning tasks rather than expending resources on processing lengthy contexts.

% \textbf{Learnability and Simplicity}

% Finally, we prioritize data entries that are straightforward and easy to interpret. Complex or domain-specific code requiring extensive background knowledge is excluded. This ensures that the dataset remains accessible for general-purpose models.

% \textbf{Rationale}: Simplifying the dataset reduces cognitive load during model training and inference, enabling faster learning and better generalization across diverse tasks.

\begin{table*}[!ht]
    \centering
\resizebox{0.9\textwidth}{!}{
    \begin{tabular}{c|c|ccc|ccccc}
    \toprule
    
        \multirow{2}{*}{\textbf{Type}} & \multirow{2}{*}{\textbf{Datasets}} & \multicolumn{3}{c|}{\textbf{Qwen2.5-7B}}& \multicolumn{5}{c}{\textbf{Qwen2.5-Coder-7B}} \\
        
        &&\textbf{Base} & \textbf{Instruct} & \textbf{Ours} & \textbf{Base} & \textbf{Instruct} & \textbf{Ours} & \textbf{CodeI/O} & \textbf{Code I/O++} \\ 
        \midrule
        \multirow{4}{*}{\textbf{Math}} & GSM8K & 85.4 & \textbf{91.6} & \colorbox{red!20}{89.7} & 83.9 & 84.9 & \colorbox{red!20}{\textbf{89.7}} & 79.3 & 80.5\\ 
        \textbf{} & MATH & 49.8 & 74.3 & \colorbox{red!20}{\textbf{75.2}} & 46.6 & 68.0 & \colorbox{red!20}{\textbf{74.3}} & 39.8 & 39.9 \\ 
        \textbf{} & AIME24 & - & \textbf{12.3} & \colorbox{red!20}{11.7} & - & 5.8 & \colorbox{red!20}{\textbf{10.2}} & - & - \\ 
        \textbf{} & AIME25 & - & \textbf{8.5} & \colorbox{red!20}{5.6} & - & 3.1 & \colorbox{red!20}{\textbf{5.0}} & - & - \\ 
        \midrule
        \multirow{4}{*}{\textbf{Knowledge}} & GPQA & 32.5 & 34.8 & \colorbox{red!20}{\textbf{35.9}} & 30.3 & \textbf{32.1} & \colorbox{red!20}{31.0} & - & - \\ 
        \textbf{} & MMLU-PRO & 45.0 & \textbf{56.4} & \colorbox{red!20}{53.6} & 40.2 & 41.2 & \colorbox{red!20}{\textbf{48.7}} & ~ & - \\ 
        \textbf{} & WinoGrad & \textbf{73.3} & 70.5 & 73.0 & 68.2 & \textbf{70.2} & 68.0 & 68.1 & 67.6 \\ 
         \textbf{} & MMLU-Redux & 71.1 & \textbf{76.0} & \colorbox{red!20}{73.9} & 66.6 & 65.1 & \colorbox{red!20}{\textbf{69.0}}& - & - \\ 
        \midrule
        \multirow{5}{*}{\textbf{Code}} & HumanEval & 61.0 & \textbf{80.5} & \colorbox{red!20}{72.6} & 65.2 & \textbf{84.8} & \colorbox{red!20}{82.3} & - & - \\ 
       \textbf{} & MBPP & 56.1 & \textbf{73.9} & \colorbox{red!20}{59.9} & 57.9 & \textbf{76.7 }& \colorbox{red!20}{70.9} & - & - \\ 
    \textbf{} & LCB-G & 28.5 & \textbf{37.0} & \colorbox{red!20}{36.8} & 37.0  & \textbf{43.1} & 35.0 & - & - \\ 
        \textbf{} & LCB-O & 40.3 & 42.3 & \colorbox{red!20}{\textbf{48.2}} & 4.0 & \textbf{9.1} & \colorbox{red!20}{8.14} & 17.6 & 20.6 \\
        \textbf{} & CRUX & 46.5 & 52.8 & \colorbox{red!20}{\textbf{59.8}} & 52.8 & 60.4 & \colorbox{red!20}{\textbf{63.5}} & 61.2 & 60.3\\
        \midrule
        \multirow{4}{*}{\textbf{Logic}} & DROP & 52.3 & \textbf{80.7} & \colorbox{red!20}{68.2} & 39.9 & \textbf{75.9} & \colorbox{red!20}{58.6} & 59.4 & 54.9 \\ 
        \textbf{} & BBH & 56.0 & \textbf{81.3} & \colorbox{red!20}{71.6} & 61.5 & 69.5 &\colorbox{red!20}{\textbf{ 72.9}} & 65.8 & 66.0 \\ 
        \textbf{} & ZebraLogic & 9.4 & 8.8 & \colorbox{red!20}{\textbf{10.8}} & 9.2 & \textbf{10.7} & 8.8 & 8.6 & 9.3 \\ 
        \textbf{} & KOR-Bench & 33.4 & \textbf{45.1} & \colorbox{red!20}{36.4} & 30.2 & 32.0 & \colorbox{red!20}{32.1} & \textbf{43.4} & 41.8 \\ \bottomrule
    \end{tabular}
    }
    \caption{Performance of the TeaR model initialized with the base model on four domains. The best results within the same model series are \textbf{bolded}. Our results that show improvement over the base model are  marked in \colorbox{red!20}{red}.}
    \label{tab:main}
    \vspace{-0.5em}
\end{table*}

\section{Experiments}
In this section, we conduct extensive experiments to evaluate TeaR, guided by the following research questions: \textbf{- RQ1} How does the performance of TeaR vary when initialized with different base models?  \textbf{- RQ2} How does TeaR perform in long-CoT scenarios? \textbf{- RQ3} How does the performance of TeaR compare with that of other data composition strategies? \textbf{- RQ4} Which reasoning capabilities benefit most from TeaR?
% \begin{itemize}[leftmargin=*]
%     \item \textbf{RQ1} How does the performance of TeaR vary when initialized with different base models?
%     \item \textbf{RQ2} How does TeaR perform in long-COT scenarios?
% \item \textbf{RQ3} How does the performance of TeaR compare with that of other data composition strategies?
% \item \textbf{RQ4} Which reasoning capabilities benefit most from TeaR?
% \item \textbf{RQ5} vs Code RL

%\end{itemize}

\paragraph{Evaluation Dataset}

To comprehensively assess the reasoning capabilities of TeaR-enhanced models and strong baselines, we evaluate across 17 benchmarks spanning four major reasoning domains: \textbf{mathmatics}, \textbf{logical reasoning}, \textbf{general knowledge}, \textbf{code reasoning}. 
For mathematical reasoning, we select 5 diverse datasets: GSM8K~\citep{cobbe2021trainingverifierssolvemath}, MATH~\citep{hendrycks2021measuringmathematicalproblemsolving}, AMC, AIME24, and AIME 25~\citep{aime_1983_2024}.
For general knowledge, we select MMLU-Pro~\cite{wang2024mmlu}
, MMLU-Redux~\cite{gema2024we}, GPQA~\cite{rein2024gpqa}, and WinoGrad~\cite{sakaguchi2021winogrande}. 
For logical reasoning, we include BBH~\cite{suzgun-etal-2023-challenging}, Zebra Logic~\citep{lin2025zebralogicscalinglimitsllms}, DROP~\citep{dua-etal-2019-drop}, and KORBench~\citep{ma2025korbenchbenchmarkinglanguagemodels}.
For code reasoning, we evaluate on HumanEval~\cite{chen2021evaluating}, MBPP~\cite{austin2021program}, LiveCodeBench~\citep{jain2024livecodebench} and CRUXEval \cite{pmlr-v235-gu24c} .
See Appendix \ref{bench_info} for details of evaluation datasets.

\paragraph{Implementation Deatils.}   
To ensure the generality and robustness of our findings, we select two representative base models for evaluation: Qwen-2.5-7B and Qwen-2.5-Coder-7B~\cite{qwen2025qwen25technicalreport}. We further select three distilled models to validate the effectiveness of our approach on inference models: R1-Distilled-1.5B, R1-Distilled-7B, and R1-Distilled-32B~\cite{guo2025deepseek}. These models cover a range of architectures, including general-purpose and code-specialized variants, as well as long-chain-of-thought scenarios, enabling a comprehensive assessment of the effectiveness of TeaR.

For all models trained from the base model, we use a batch size of 256, rollout length of 8, temperature of 1.1, learning rate of 1e-6, and a maximum token length of 4,096 during training. Throughout the training process, we use the MATH-test set as the validation dataset to help us select appropriate checkpoints.  
For models trained starting from the distilled model, we keep the batch size, rollout length, temperature, and learning rate consistent with those used for the base model. However, we increase the maximum token length to 16,384. All our training is conducted using the veRL.

\begin{table*}[!ht]
    \centering
    \resizebox{0.95\textwidth}{!}{
    \begin{tabular}{l|cccccccccccccccc}
    \toprule
    \textbf{Model} & \textbf{AMC} & \textbf{AIME24} & \textbf{AIME25} & \textbf{GPQA} & \textbf{M.PRO} & \textbf{M.Redux} & \textbf{LCB} & \textbf{CRUX} & \textbf{DROP} & \textbf{ZebraLogic} & \textbf{KOR-Bench} & \textbf{Average} \\
    \midrule
    Distill-1.5B & 71.7 & 29.0 & 24.6 & 36.9 & 27.1 & 34.0 & 17.4 & 34.3 & 44.2 & 5.1 & 11.0 & 30.5 \\
    \textbf{Ours} & \textbf{72.0} & 28.7 & \textbf{26.5} & 35.9 & \textbf{28.1} & \textbf{37.0} & \textbf{18.3} & \textbf{37.2} & \textbf{46.2} & \textbf{7.2} & \textbf{14.1} & \textbf{31.9} \\
    \midrule
    Distill-7B & 89.8 & 55.0 & 40.2 & \textbf{51.6} & 36.9 & 64.9 & 39.5 & 65.4 & 76.6 & 23.3 & 27.9 & 52.2 \\
     \textbf{Ours} & \textbf{91.4} & \textbf{58.1} & \textbf{42.3} & 51.3 & \textbf{38.5} & \textbf{68.1} & \textbf{44.3} & \textbf{69.5} & \textbf{78.8} & \textbf{30.5} & \textbf{35.2} & \textbf{55.3} \\
    \midrule
    Distill-32B & 88.8 & 71.7 & 58.3 & 61.8 & 74.0 & 87.3 & 59.1 & 86.0 & 89.0 & 67.4 & \textbf{52.9} & 72.4 \\
    \textbf{Ours} & \textbf{95.9} & \textbf{74.0} & \textbf{59.6} & \textbf{64.5} & \textbf{75.7} & \textbf{88.0} & \textbf{61.1} & \textbf{89.4} & \textbf{89.5} & \textbf{70.0} & 50.3 & \textbf{74.4} \\
    \bottomrule
    \end{tabular}
    }
    \caption{Performance on long-cot distilled models, each row represents a model variant and each column represents a dataset. The performance improvements are \textbf{bolded}. M. refers to MMLU.}
    \label{tab:longcot}
    \vspace{-0.7em}
\end{table*}

\subsection{Main Performance}
Table~\ref{tab:main} shows a comparison between our model, trained from the base model, and the corresponding instruct models in the same series. Moreover, we also compare with two strong baselines that use code for improving reasoning ability: (1) CodeI/O , which comprises 3.5M input-output code samples paired with final-answer supervision; and (2) CodeI/O++ , a lightly augmented variant featuring minimal structural enhancements\footnote{We report the results of CodeI/O + Tulusft to avoid the influence of its in-house instruction tuning datasets.}.  Our main conclusions are as follows:
\begin{itemize}[leftmargin=*]
    \item \textbf{vs. Base Model} The consistent performance improvements across almost all benchmarks over the base model demonstrate the effectiveness of our approach. Notably, significant gains are observed on datasets that require strong reasoning capabilities, such as Math, Logic, and Code. For instance, the accuracy on Math increases from 49.8\% to 75.2\%, on DROP from 52.3\% to 68.2\%, and on BBH from 56.0\% to 71.6\%. Encouragingly, even on tasks like GPQA and MMLU-PRO, which demand extensive domain-specific knowledge, our method still achieves notable improvements. This suggests that enhancing the model's reasoning abilities can lead to comprehensive performance gains, highlighting the importance of advancing reasoning capabilities in LLM.
    \item \textbf{vs. Intruct Model} Although we only utilized a limited set of code-related tasks during training (2.4K distinct seed questions, 200K  vs 3.5M training samples), our model still achieves comparable performance on mathematical reasoning and knowledge benchmarks, and even achieves superior results on several code-related tasks. Remarkably, on the qwen2.5-coder-7b-base backbone, our model outperforms the qwen2.5-conder-7b-instruct variant on MMLU-PRO and MMLU Redux. However, our model still lags behind the instruction-tuned counterpart on certain logic reasoning benchmarks, which can be largely attributed to the lack of explicit instruction-following tuning and the relatively niche nature of these tasks, which often require specialized optimization (e.g., KOR-bench).
    \item \textbf{7B vs. Coder-7B} When comparing the Qwen2.5-7B-Base series of models with the Qwen2.5-Coder-7B series, we observe that while the code-specialized models demonstrate significant improvements in code generation tasks (e.g., HumanEval, MBPP, and LCB-G), they suffer notable performance drops on code-related reasoning tasks (LCB-O) and logic reasoning benchmarks. This finding indirectly supports our hypothesis that learning code patterns may primarily capture syntactic structures in code generation, rather than deeper reasoning capabilities. It also motivates us to place greater emphasis on reasoning-oriented tasks in future work.
    \item \textbf{vs. CodeIO} Although our approach utilizes an order of magnitude less data compared to CodeI/O, along with fewer distinct code-related queries, we still achieve consistent improvements on five out of all ten comparable datasets, obtain comparable performance on three, and underperform on only two — where the advantage of CodeI/O may stem from its instruction fine-tuning. This further suggests that leveraging reinforcement learning to enhance the model’s reasoning capabilities, rather than merely fitting static code patterns, is a more effective way to improve generalization across diverse reasoning tasks.
\end{itemize}

% \texbtf{RQ6} Other Analysis e.g. improve difficulty add query (补一个)
\subsection{Long-CoT Performance}
While we have already demonstrated the effectiveness of TeaR when trained directly from a base model, most existing reasoning models are typically built upon Long-CoT (Chain-of-Thought) frameworks. To further validate our approach in such settings, we conduct additional experiments using the DeepSeek-R1-Distilled series as the starting point. The results are summarized in Table~\ref{tab:longcot}, and our key findings are as follows:
\begin{itemize}[leftmargin=*]
    \item Overall, TeaR demonstrates improvements over its starting-point models across three model scales, further validating its effectiveness in the Long-CoT setting. A closer examination reveals that performance gains are relatively modest on tasks such as Math and code-related benchmarks, while significant improvements are consistently observed on general-purpose and logic reasoning datasets. This may be attributed to the fact that the distilled models used as starting points have already been heavily optimized for mathematical and code generation tasks.
    \item When comparing performance across models of different scales, we observe that several tasks exhibit significant improvements only at the 32B scale (e.g., AMC, GPQA). We hypothesize that this may be due to the fact that larger models generally possess stronger generalization capabilities, enabling them to better learn and apply general reasoning strategies from reasoning tasks. Furthermore, the distilled-1.5B model shows relatively minor improvements across the board, which may further support our hypothesis from an indirect perspective.
\end{itemize}

\begin{table}[t]
    \centering
    \scalebox{0.85}{
    \begin{tabular}{l|cccc}
    \toprule
        & \textbf{Math} & \textbf{General} & \textbf{Code} & \textbf{Logic}  \\ 
        \midrule
        {Base} & 63.3 & 56.5 & 40.6 & 37.7 \\ 
        
        {Math } & 82.5 & 58.0 & 55.0 & 43.2 \\
        {w/ Ref Code } & 82.4 & 57.9 & 54.6 & 42.3
        \\
        \textbf{Ours} & \cellcolor{red!20} 83.1 & \cellcolor{red!20} 59.1 & \cellcolor{red!20} 55.5 & \cellcolor{red!20} 46.7 \\ 
        \midrule
        {Base} & 61.7 & 51.3 & 52.5 & 42.6 \\ 
        Code & 62.7 & 52.5 & 43.5 & 44.3 \\ 
         \textbf{Ours} & \cellcolor{red!20} 63.9 & \cellcolor{red!20}52.6 & \cellcolor{red!20}56.9 & \cellcolor{red!20}48.2 \\ \bottomrule
    \end{tabular}
    }
    \caption{Data source (Math and Code) and format (w/ Ref Code) comparsion. Best results are in \colorbox{red!20}{red}.}
    \label{tab:ablation}
    \vspace{-0.5em}
\end{table}
\subsection{Data Analysis}
% To further investigate 使用代码相关的推理属性任务的优越性 we also compare the following data format，训练方式与我们保持一致: (1) Math: 我们使用Math训练集直接参与训练，Reward 采用veRL自带的math verifier (2) With ref code: 在训练的过程中与codeI/O类似，在query中添加reference code要求模型follow code 的逻辑进行推理 (3) Raw code: 直接要求模型生成相应问题的代码，对于每个代码我们使用在sandbox 环境中使用所有的测试样例来判断正确性，如果通过所有测试样例则reward 为1 否则为0。对于(1)和(2)我们从base模型直接训练，对于(3)由于training from qwen2.5-7b-base训练reward 长期为0，这可能是由于base模型相应的指令遵循能力弱的问题，因此我们从R1-distilled-7B开始训练，我们的结果在Table~\ref{}中展示，主要结论如下
To further investigate the advantages of our approach in leveraging code-related reasoning attributes, we also compare several alternative data formats under the same training paradigm. Specifically, we consider the following three settings:
(1) Math : We directly use the Math training set for model fine-tuning. The reward signal is provided by the built-in math verifier from veRL.
(2) With Ref Code : Similar to the codeI/O setting, we include a reference code snippet in the input query, instructing the model to follow the logic of the given code during reasoning.
(3) Code : Asking the mode to generate code directly for the same problem. For each generated code, we execute it in a sandbox environment and evaluate its correctness based on whether it passes all test cases. A reward of 1 is assigned if all tests are passed; otherwise, it is 0.
For settings (1) and (2), we initialize training from the Qwen2.5-7B-base model. However, for setting (3), we observe that the reward remains consistently zero when starting from Qwen2.5-7B-base, likely due to the weak instruction-following capability of the base model. Therefore, we instead initialize training from R1-distilled-7B. The results are summarized in Table~\ref{tab:ablation}, from which we draw the following main conclusions:
\begin{itemize}[leftmargin=*]
    \item By comparing TeaR with Math and Ref Code, we observe that TeaR demonstrates superior performance on almost all benchmarks, particularly in logical reasoning tasks, where the most significant improvements highlight the generalization and robustness of our method in enhancing reasoning capabilities. Furthermore, when comparing Math and With Ref Code, the overall performance of With Ref Code is relatively weaker, further supporting our hypothesis that relying entirely on code-based guidance may lead the model to overfit suboptimal reasoning patterns.
    \item Compared to Raw Code, we observe that training the R1-Distill-7B model using code generation with RL brings almost no performance gain. We attribute this to the fact that the distilled model has already been heavily optimized for code generation tasks, and the relatively low difficulty of our training data limits its ability to further improve the model’s capabilities. This also suggests a lack of general reasoning ability in the model. Moreover, although training with raw code provides little benefit, our method is still able to improve code generation performance, further indicating that the reasoning related task is more important than the code itself.
\end{itemize}
% 对比TeaR 和 Math 和 With Ref Code，我们可以看到在几乎所有测试上TeaR都展示了其优越性，尤其是在逻辑推理任务上，提升空间最大充分说明了我们的方法对于reasoning能力的泛化性和鲁棒性。Besides，对比Math 和 With Ref Code，我们可以看到With Ref Code的整体性能要弱上一截， further verify 了 完全使用代码辅导可能会让模型在suboptimal的reasoning pattern上过拟合的assumption 

\begin{table}[t]
    \centering
    \scalebox{0.75}{
    \begin{tabular}{l|ccccc}
    \toprule
        & Logic & Tracking & Object & Movie  \\
        & Deduction & Objects & Counts & Rec.\\
        \midrule
        {Base} & 64.8 & 76.0 & 22.8 & 69.6 \\ 
        \textbf{Ours} & \cellcolor{red!20} 88.0 & \cellcolor{red!20} 88.8 &\cellcolor{red!20}  58.4 & \cellcolor{red!20} 85.6 \\ 
        \midrule
        {Coder-Base} & 76.0 & 66.8 & 62.0 & 80.0 \\ 
        \textbf{Ours} & \cellcolor{red!20} 88.0 & \cellcolor{red!20} 90.4 & \cellcolor{red!20} 69.2 & \cellcolor{red!20} 86.8 \\ 
     \bottomrule
    \end{tabular}
    }
    \caption{Comparaison between Qwen2.5-7B-base and Qwen2.5-Coder-7B-base under BBH subsets. Best results are in \colorbox{red!20}{red}.}
    \label{tab:bbh}
    \vspace{-0.5em}
\end{table}
\subsection{Fine-grained Analysis}
To gain deeper insight into the qualitative improvements brought by TeaR, we conduct a fine-grained analysis on several challenging reasoning subsets from the BBH benchmark.  We carefully selected the following four tasks that are strongly related to reasoning abilities. These tasks evaluate different aspects of logical and compositional reasoning: (1) \textit{Tracking Shuffled Objects}: It requires the model to accurately track sequential transformations of objects, testing its ability to perform deductive reasoning over dynamic object states. (2) \textit{Object Counting}: A reasoning task that evaluates a model's ability to count objects and integrate results based on the given context and question. (3) \textit{Logical Deduction}: This task assesses the model’s capability to perform deductive reasoning by interpreting clues and understanding spatial relationships and placements among entities. (4) \textit{Movie Recommendation}: Given a context of user behaviors and require the model to predict the movies that user might like according to his/her history interactions. This task demonstrates the generalization of the model's reasoning capabilities — whether our approach can yield further improvements on tasks that are seemingly unrelated. 

As shown in Table~\ref{tab:bbh}, our method achieves significant improvements across all four subsets, especially on logic deduction and object counting tasks, where the performance is improved by 35.8\% and 156.1\%, respectively, over the Qwen2.5-7B-Base model. These results strongly indicate that our method effectively enhances the model's overall reasoning ability. Even on the seemingly unrelated movie recommendation task, the improved reasoning capability leads to consistent gains on both base models, further supporting the effectiveness and generality of our approach.

\section{Related Work}
\paragraph{Reasoning Capabilities of LLMs}

Reasoning ability has long been a core capability of LLMs. Research has shown that as model scale increases, reasoning capabilities naturally emerge as a fundamental ability of LLMs~\cite{DBLP:journals/tmlr/WeiTBRZBYBZMCHVLDF22}. To enhance this capability, researchers have employed various strategies: early efforts focused primarily on prompt-based methods~\cite{wei2022chain,DBLP:conf/iclr/0002WSLCNCZ23,DBLP:conf/emnlp/0001GHW00023}.
% such as Chain-of-Thought (CoT)~\cite{wei2022chain}, self-consistency~\cite{DBLP:conf/iclr/0002WSLCNCZ23}, and self-improvement~\cite{DBLP:conf/emnlp/0001GHW00023}. 
Recently, training-based approaches to improve reasoning have become mainstream, where SFT~\cite{DBLP:journals/corr/abs-2308-10792} enables models to learn advanced reasoning patterns from human-annotated or distilled data~\cite{DBLP:journals/corr/abs-2411-15382}, while Reinforcement Learning (RL)~\cite{DBLP:journals/corr/abs-2501-17161} further promotes generalization and self-learning through implicit feedback. Notably, RL with verifiable rewards has advanced reasoning capabilities significantly, particularly in mathematics and programming~\cite{DBLP:journals/corr/abs-2411-15124,shao2024deepseekmath}. However, such tasks typically concentrate on a limited set of specific domains, with little attention paid to the holistic improvement of the model's general reasoning abilities. 
In contrast to these works, TeaR aims to enhance models' general reasoning capabilities.
%In the field of RL, Proximal Policy Optimization (PPO) , which alternates between environment interaction for data collection and random gradient ascent for optimizing surrogate objective functions, has demonstrated superior performance compared to Direct Preference Optimization (DPO). Subsequently, ReMax combined variance reduction and REINFORCE techniques to eliminate the need for additional value models in PPO. Building upon this, DeepSeekMath introduced Group Relative Policy Optimization (GRPO), which replaces traditional value models with improved sampling strategies, significantly accelerating learning efficiency and achieving mathematical performance comparable to GPT-4. In contrast to these works, our proposed TeaR paradigm aims to enhance models' general reasoning capabilities rather than focusing on specific domains.

% with researchers utilizing models to simulate code execution processes before the era of LLMs
\paragraph{Leveraging Code Execution}
Code execution has long been explored in reasoning tasks~\cite{DBLP:conf/acl/JayannavarNH20,DBLP:journals/corr/ZarembaS14}. With the advent of LLMs, research focus has diverged: some works concentrate on code output prediction itself~\cite{DBLP:conf/icse/DingSPKLR24}, while others aim to enhance model's code generation capabilities through final feedback~\cite{DBLP:journals/pacmpl/DingMKR24} or intermediate paths~\cite{DBLP:conf/nips/DingPMKYR24}, which has also given rise to benchmarks for testing code generation tasks, such as CTRUXEval~\cite{DBLP:conf/icml/GuRLSS024} and LiveCodeBench~\cite{DBLP:conf/iclr/JainHGLYZWSSS25}. Recently, CodeI/O~\cite{li2025codeiocondensingreasoningpatterns} has improved model reasoning capabilities through code-referenced rejection sampling, requiring models to infer test case outputs based on code reasoning patterns. However, this approach has limitations: code is often designed with large-scale data, and not all reasoning tasks require complex algorithmic design or intricate code logic. Moreover, a complete reliance on code-oriented reasoning patterns may introduce unnecessary computational overhead, especially for simpler problems. Motivated by these observations, we propose TeaR, which aligns more closely with natural human reasoning patterns. TeaR enables models to autonomously explore effective reasoning strategies for algorithmic tasks, thereby enhancing their general reasoning capabilities in a more flexible and efficient manner.
\section{Conclusion}
In this paper, we focus directly on the algorithmic task itself and propose TeaR, a novel framework that encourages models to discover effective reasoning paths autonomously, without relying on any code references. This approach aims to enhance the model's generalization capabilities through intrinsic reasoning. We conduct experiments on both base models and long-CoT distilled models, evaluating TeaR across 17 diverse benchmark tasks. Our results demonstrate that TeaR significantly improves performance on both in-domain and cross-domain reasoning tasks, and even generalizes well to user behavior reasoning. Further analysis confirms its strong generalization ability and effectiveness. In summary, TeaR offers a fresh perspective on leveraging code-related problems to improve LLMs’ reasoning capabilities holistically, providing new insights into the development of more robust and versatile reasoning LLMs.
\section*{Limitations}
In this study, we leveraged datasets primarily sourced from existing algorithm competition platforms such as LeetCode. While these datasets provide a structured collection of algorithmic challenges, they inherently possess limitations in terms of complexity and diversity; the difficulty level may not pose significant challenges to advanced problem-solving LLMs. Furthermore, the current scale of these datasets is relatively modest, suggesting ample room for further exploration and expansion. This limitation may impact the generalizability and robustness of our findings across different problem domains.
Additionally, in this work, our backbone models were restricted to a maximum of 32B parameters. Consequently, whether larger models can sustain the conclusions we have drawn remains uncertain and untested. Exploring the performance and scalability of more extensive architectures could provide valuable insights into the adaptability of our methodology to increased model complexity.

\section*{Ethic Statement}
In this work, we are committed to upholding the standards of ethical data collection and analysis, implementing robust safeguards to ensure fairness, transparency, and scholarly integrity throughout all phases of investigation. We explicitly acknowledge our responsibility as researchers to conduct methodologically sound and ethically conscious work that advances knowledge. Recognizing the far-reaching societal implications of our research findings, we steadfastly adhere to established ethical guidelines while maintaining the standards of academic rigor. Through meticulous documentation of data quality, collection procedures, and analytical methods, we aim to ensure reproducibility and accountability in our scholarly work. We remain cognizant of emerging ethical considerations in contemporary research practices and are committed to adapting our protocols as necessary to meet evolving standards in the field, while consistently upholding the fundamental principles of research integrity and scientific excellence.

\newpage
\bibliography{custom}

\appendix
\section{Benchmark Information}
\label{bench_info}

To comprehensively assess the reasoning capabilities of TeaR-enhanced models and strong baselines, we evaluate across 17 benchmarks spanning four major reasoning domains: \textbf{mathmatics}, \textbf{logical reasoning}, \textbf{general knowledge}, \textbf{code reasoning}. 

\subsection{Mathmatics}
\paragraph{GSM8K}~\citep{cobbe2021trainingverifierssolvemath} A comprehensive dataset of 1,000 elementary-level word problems designed to assess students' multi-step mathematical reasoning and problem-solving abilities through practical scenarios.

\paragraph{MATH}~\citep{hendrycks2021measuringmathematicalproblemsolving} A diverse collection of 12,500 advanced mathematics competition problems spanning multiple disciplines to evaluate sophisticated mathematical reasoning and problem-solving capabilities.

\paragraph{AMC, AIME24 and AIME25}~\citep{aime_1983_2024} These datasets contain mathematical problems from two prestigious American mathematics competitions: the American Mathematics Competition (AMC) and the American Invitational Mathematics Examination (AIME). While the AMC serves as the initial competition open to all high school students, the AIME is a more advanced level that requires students to first excel in the AMC. Both competitions are highly respected in the United States and Canada, with AIME representing a particularly challenging level of secondary school mathematics competition that only top-performing students qualify to participate in.
\subsection{General Knowledge}
\paragraph{MMLU-Pro}~\cite{wang2024mmlu} MMLU-Pro is an enhanced version of the MMLU benchmark that features more challenging reasoning-focused questions, expanded answer options (from 4 to 10), and improved stability across different prompting methods, designed to better differentiate between advanced language models' capabilities.

\paragraph{MMLU-Redux}~\cite{gema2024we} MMLU-Redux is a carefully re-annotated subset of 5,700 questions from the original MMLU benchmark, created to address and correct the significant number of ground truth errors found in the original dataset, providing a more accurate assessment tool for evaluating language models' capabilities.

\paragraph{GPQA}~\cite{rein2024gpqa} GPQA is a challenging dataset of 448 expert-crafted multiple-choice questions in biology, physics, and chemistry, designed to be "Google-proof" and difficult for both skilled non-experts and AI systems, aiming to test and develop methods for human experts to effectively oversee AI systems that may surpass human capabilities.

\paragraph{WinoGrad}~\cite{sakaguchi2021winogrande} The Winograd Schema Challenge (WSC) is a benchmark of 273 expert-created pronoun resolution problems designed to test commonsense reasoning abilities.

\subsection{Logical Reasoning}
\paragraph{BBH}~\cite{suzgun-etal-2023-challenging} BBH (Beyond the Imitation Game Benchmark) is a collection of 23 complex reasoning tasks that were selected from the BIG-Bench collection. These tasks were specifically chosen because they present significant challenges for LLMs. The benchmark covers diverse types of reasoning, including logical thinking, mathematical problem-solving, and procedural reasoning abilities.

\paragraph{ZebraLogic}~\cite{lin2025zebralogicscalinglimitsllms} ZebraLogic is an evaluation framework that uses logic grid puzzles with controllable complexity to systematically assess and understand the logical reasoning capabilities and limitations of LLMs, revealing a "curse of complexity" where model performance significantly declines as puzzle difficulty increases.

\paragraph{DROP}~\cite{dua-etal-2019-drop} DROP (Discrete Reasoning Over Paragraphs) is a challenging reading comprehension benchmark containing 96,000 questions that require systems to perform discrete operations (like counting, sorting, and arithmetic) while understanding textual content, designed to test more comprehensive comprehension abilities than previous datasets.

\paragraph{KORBench}~\cite{ma2025korbenchbenchmarkinglanguagemodels} KorBench is a specialized evaluation tool created to measure a model's fundamental reasoning and planning capabilities while trying to reduce the impact of knowledge gained during pre-training. The benchmark consists of five different categories (Operation, Logic, Cipher, Puzzle, and Counterfactual), with each category containing 25 unique, hand-crafted rules that were specifically created for this purpose. By using these novel rules rather than relying on common knowledge, KorBench can better assess how well models can adapt to and understand new rule-based challenges they haven't encountered before.
\subsection{Code Reasoning}
\paragraph{HumanEval}~\cite{chen2021evaluating} HumanEval is an evaluation dataset designed to test code generation models' ability to write functionally correct Python programs based on docstring descriptions, measuring how well models can translate natural language specifications into working code.

\paragraph{MBPP}~\cite{austin2021program} MBPP (Mostly Basic Programming Problems) is a dataset containing 974 entry-level programming tasks designed to evaluate language models' ability to generate Python code from natural language descriptions, suitable for testing basic programming problem-solving capabilities.

\paragraph{LiveCodeBench}~\cite{jain2024livecodebench} LiveCodeBench is a continuously updated, contamination-free benchmark that collects recent programming problems from LeetCode, AtCoder, and CodeForces, designed to evaluate a broader range of coding capabilities in language models, including code generation, self-repair, code execution, and test output prediction.

\paragraph{CRUXEval}~\cite{pmlr-v235-gu24c} CRUXEval is a benchmark that tests how well AI models can understand and process code by challenging them to predict either the inputs or outputs of Python functions where the names have been anonymized. The benchmark tests the model's ability to mentally execute code and track how variables change throughout the program's execution. This involves understanding how the program flows and maintaining awareness of the program's state at different stages of execution.

\section{Case Study}
\label{app:case}
This section shows an example of TeaR.

\begin{exmp}{An Example of Query}{exmp:query}

Given an integer array A of length n, with initial values provided by the user. You need to handle the following two types of operations:

\textbf{Query Operation}

Format: `query X Y`  

Ask for the maximum value in array A from index X to Y (inclusive of both ends).

\textbf{Add Operation} 

Format: `add X Y Z`  

Increase all elements in array A from index X to Y (inclusive of both ends) by Z.

\textbf{Input:}

A = [1, 2, 3]  

add 1 2 3  

add 2 3 -1  

add 1 3 2  

query 1 3

\end{exmp}

% 设置代码样式
\definecolor{codegreen}{rgb}{0,0.6,0}
\definecolor{codegray}{rgb}{0.5,0.5,0.5}
\definecolor{codepurple}{rgb}{0.5,0,0.5}
\definecolor{backcolour}{rgb}{0.95,0.95,0.92}

\lstdefinestyle{mystyle}{
    backgroundcolor=\color{backcolour},   
    commentstyle=\color{codegreen},
    keywordstyle=\color{magenta},
    numberstyle=\tiny\color{codegray},
    stringstyle=\color{codepurple},
    basicstyle=\ttfamily\footnotesize,
    breakatwhitespace=false,         
    breaklines=true,                 
    captionpos=b,                    
    keepspaces=true,                 
    numbers=left,                    
    numbersep=5pt,                  
    showspaces=false,                
    showstringspaces=false,
    showtabs=false,                  
    tabsize=2
}

\lstset{style=mystyle}

% 插入代码示例
\begin{lstlisting}[language=Python, caption=Example Python Code]
class SegmentTree:
    def __init__(self, data):
        self.n = len(data)
        self.tree = [0] * (4 * self.n)  
        self.lazy = [0] * (4 * self.n)  
        self.build(1, 0, self.n - 1, data)

    def build(self, node, start, end, data):
        if start == end:
            self.tree[node] = data[start]
        else:
            mid = (start + end) // 2
            left_child = 2 * node
            right_child = 2 * node + 1
            self.build(left_child, start, mid, data)
            self.build(right_child, mid + 1, end, data)
            self.tree[node] = max(self.tree[left_child], self.tree[right_child])

    def push_down(self, node, start, end):
        if self.lazy[node] != 0:
            mid = (start + end) // 2
            left_child = 2 * node
            right_child = 2 * node + 1
            self.tree[left_child] += self.lazy[node]
            self.lazy[left_child] += self.lazy[node]
            self.tree[right_child] += self.lazy[node]
            self.lazy[right_child] += self.lazy[node]
            self.lazy[node] = 0

    def update(self, node, start, end, l, r, val):
        if r < start or end < l:
            return
        if l <= start and end <= r:
            self.tree[node] += val
            self.lazy[node] += val
            return
        self.push_down(node, start, end)
        mid = (start + end) // 2
        left_child = 2 * node
        right_child = 2 * node + 1
        self.update(left_child, start, mid, l, r, val)
        self.update(right_child, mid + 1, end, l, r, val)
        self.tree[node] = max(self.tree[left_child], self.tree[right_child])

    def query(self, node, start, end, l, r):
        if r < start or end < l:
            return float('-inf')
        if l <= start and end <= r:
            return self.tree[node]
        self.push_down(node, start, end)
        mid = (start + end) // 2
        left_child = 2 * node
        right_child = 2 * node + 1
        left_max = self.query(left_child, start, mid, l, r)
        right_max = self.query(right_child, mid + 1, end, l, r)
        return max(left_max, right_max)


n, q = map(int, input().split())
data = list(map(int, input().split()))
st = SegmentTree(data)

for _ in range(q):
    cmd, *args = input().split()
    if cmd == "query":
        x, y = map(int, args)
        print(st.query(1, 0, n - 1, x - 1, y - 1))
    elif cmd == "add":
        x, y, z = map(int, args)
        st.update(1, 0, n - 1, x - 1, y - 1, z)
\end{lstlisting}
\begin{exmp}{An Example of Output with Referenced Code}{exmp:output-with}
To solve the problem listed in the given input using the reference code snippet, let's step through each operation and understand how they affect the segment tree structure:

1. \textbf{Build Initial Segment Tree:}  

   Initial array A = [1, 2, 3]  
   
   The segment tree is built to represent the maximum in various segments of the array. Initially, without any operations, the tree represents:
   
   - Range (1, 1) -> 1
   
   - Range (1, 2) -> max(1, 2) = 2
   
   - Range (1, 3) -> max(1, 2, 3) = 3
   
   - Range (2, 2) -> 2

   - Range (2, 3) -> max(2, 3) = 3
   
   - Range (3, 3) -> 3

2.  \textbf{First Operation} - `add 1 2 3`:  

   Increases elements from index 1 to 2 by 3:  
   
   - A becomes [4, 5, 3]  
   
   Modify the segment tree with updates in range:
   
   - Range (1, 1) -> 4

   - Range (1, 2) -> max(4, 5) = 5
   
   - Range (1, 3) -> max(4, 5, 3) = 5
   
   - Range (2, 2) -> 5
   
   - Range (2, 3) -> max(5, 3) = 5
   
   The lazy propagation ensures that these updates correctly reflect the updated maximum in these segments.

3.  \textbf{Second Operation} - `add 2 3 -1`:  
   
   Decrease elements from index 2 to 3 by 1:  
   
   - A becomes [4, 4, 2]  
   
   Update the segment tree:
   
   - Range (1, 2) remains => max(4, 4) = 4
   
   - Range (1, 3) => max(4, 4, 2) = 4
   
   - Range (2, 2) -> 4
   
   - Range (2, 3) => max(4, 2) = 4

4.  \textbf{Third Operation} - `add 1 3 2`:  

   Increase elements from index 1 to 3 by 2:  
   
   - A becomes [6, 6, 4]  
   
   Update the segment tree:
   
   - Range (1, 2) => max(6, 6) = 6
   
   - Range (1, 3) => max(6, 6, 4) = 6
   
   - Range (2, 3) => max(6, 4) = 6
   
   - Every range within 1 to 3 reflects the updated values.

5.  \textbf{Query Operation} - `query 1 3`:  

   Return the maximum value in array A from index 1 to 3: 
   
   The segment tree reflects:
   
   - Range (1, 3) has maximum value of 6.

So, the result of the `query 1 3` operation after all these updates is 6.
\end{exmp}

\begin{exmp}{An Example of Output without Referenced Code}{exmp:output-without}
To solve this sequence of operations, let's process each step one by one, modifying the array accordingly:

1. \textbf{Initial Array}: A = [1, 2, 3]

2. \textbf{Operation}: `add 1 2 3`  
   Increase elements from index 1 to 2 by 3.  
   Modified Array after operation: A = [4, 5, 3]

3. \textbf{Operation}: `add 2 3 -1`  
   Decrease elements from index 2 to 3 by 1.  
   Modified Array after operation: A = [4, 4, 2]

4. \textbf{Operation}: `add 1 3 2`  
   Increase elements from index 1 to 3 by 2.  
   Modified Array after operation: A = [6, 6, 4]

5. \textbf{Operation}: `query 1 3`  
   Find the maximum value from index 1 to 3 in the array.  
   The values to consider are 6, 6, and 4.

The maximum value in this range is 6.
\end{exmp}

\end{document}